\documentclass[conference]{IEEEtran}
\usepackage{graphicx, tabularx, tikz, xcolor}
\usepackage{amsmath}
\usepackage{listings}

\usepackage{graphicx}
\usepackage{siunitx}
\usepackage{booktabs}
\usepackage{comment}
\usepackage{multirow}
\usepackage[font=small]{caption}
\usepackage{url}
	
\usepackage{longtable}

\usepackage[capitalise,noabbrev]{cleveref}

\usepackage[utf8]{inputenc}
\usepackage[acronym]{glossaries}

\newacronym{3gppc-v2x}{3GPP C-V2X}{3GPP cellular vehicle-to-everything}
\newacronym{cqr}{CQR}{conformalized quantile regression}
\newacronym{ai}{AI}{artificial intelligence}
\newacronym{bs}{BS}{base station}
\newacronym{cdf}{CDF}{cumulative distribution function}
\newacronym{gcd}{GCD}{greatest common divisor}
\newacronym{gbqr}{GBQR}{gradient boosting quantile regression}
\newacronym{gb}{GB}{gradient boosting}
\newacronym{intel_nuc}{Intel NUC}{Intel Next Unit of Computing}
\newacronym{gps}{GPS}{global positioning system}
\newacronym{mcs}{MCS}{modulation and coding scheme}
\newacronym{ml}{ML}{machine learning}
\newacronym{mda}{MDA}{mean decrease in accuracy}
\newacronym{lcm}{LCM}{least common multiple}
\newacronym{nn}{NN}{neural network}
\newacronym{lr}{LR}{Linear regression}
\newacronym{mse}{MSE}{mean square error}
\newacronym{mae}{MAE}{mean absolute error}
\newacronym{mdi}{MDI}{mean decrease in impurity}
\newacronym{kde}{KDE}{kernel density estimate}
\newacronym{rl}{RL}{reinforcement learning}
\newacronym{rf}{RF}{random forest}
\newacronym{rsrp}{RSRP}{reference signal received power}
\newacronym{rssi}{RSSI}{received signal strength indicator}
\newacronym{rb}{RB}{resource block}
\newacronym{qos}{QoS}{quality of service}
\newacronym{sdr}{SDR}{software defined radio}
\newacronym{snr}{SNR}{signal-to-noise ratio}
\newacronym{sgd}{SGD}{stochastic gradient descent}
\newacronym{pcap}{PCAP}{packet capture}
\newacronym{srs}{SRS}{Software Radio Systems}
\newacronym{rmse}{RMSE}{root mean square error}
\newacronym{per}{PER}{packet error rate}
\newacronym{sps}{SPS}{Semi-Persistent Scheduling}

\newacronym{usrp}{USRP}{universal software radio peripheral}
\newacronym{ue}{UE}{user equipment}
\newacronym{v2x}{V2X}{vehicle-to-everything}
\newacronym{prr}{PRR}{packet reception ratio}

\IEEEoverridecommandlockouts                              

\begin{document}

\title{From Empirical Measurements to Augmented Data Rates: A Machine Learning Approach for MCS Adaptation in Sidelink Communication\\
\thanks{\textbf{\textsc{Acknowledgement:}} This work was partially supported by the Federal Ministry of Education and Research (BMBF) of the Federal Republic of Germany as part of the AI4Mobile (16KIS1170K) and 6G-RIC (16KISK020K) projects. The authors alone are responsible for the content of the paper.}
}

\author{\IEEEauthorblockN{Asif Abdullah Rokoni, Daniel Sch\"aufele,
Martin Kasparick,
and S{\l}awomir Sta\'nczak}
\IEEEauthorblockA{Wireless Communications and Networks,
Fraunhofer Heinrich Hertz Institute, Berlin, Germany\\
\{asif.abdullah.rokoni, daniel.schaeufele, martin.kasparick, slawomir.stanczak\}@hhi.fraunhofer.de}
}

\maketitle
\thispagestyle{empty}
\pagestyle{empty}

\begin{abstract}

Due to the lack of a feedback channel in the C-V2X sidelink, finding a suitable \gls{mcs} is a difficult task. However, recent use cases for \gls{v2x} communication with higher demands on data rate necessitate choosing the \gls{mcs} adaptively. In this paper, we propose a machine learning approach to predict suitable \gls{mcs} levels. Additionally, we propose the use of quantile prediction and evaluate it in combination with different algorithms for the task of predicting the \gls{mcs} level with the highest achievable data rate. Thereby, we show significant improvements over conventional methods of choosing the \gls{mcs} level. Using a machine learning approach, however, requires larger real-world data sets than are currently publicly available for research. For this reason, this paper presents a data set that was acquired in extensive drive tests, and that we make publicly available.


\end{abstract}

\glsresetall
\section{Introduction}

Recently, \gls{v2x} communication has gained a lot of attention due to its potential to increase road safety, traffic efficiency, but also to support autonomous driving technologies, for example, by extending on-board sensors with information from the infrastructure. Sidelink communication, as introduced by 3GPP in the LTE and 5G standards \cite{garcia2021tutorial}, has the advantage that it can operate both, in cellular coverage, where certain configuration is provided by the \gls{bs} (called Mode 3 in LTE V2X), and out-of-coverage, where the \glspl{ue} autonomously configure themselves (called Mode 4 in LTE V2X). Although in many of today's most common sidelink use cases only very few information messages are transmitted, more recent use cases, which involve video transmissions \cite{zhang2017effect}, require significantly higher goodput. 
Here, it is paramount to configure the sidelink according to the propagation conditions, and thus methods for choosing good transmission parameters are required \cite{ku2022adaptive}. One solution for this problem is to use \gls{ml} algorithms, which, however, requires training data. While training data for cellular communication can be found easily \cite{sliwa_empirical_2019, schaufele2021terminal, palaios2021network}, there is a lack of available training data for sidelink communications. Most of the existing literature focuses on very specific scenarios,  and their measurements are not publicly available for use in \gls{ml}-related research \cite{rodriguez2016measurement, de2021its, husges2022simulation}.

Due to the lack of sidelink measurement data, we conducted an extensive
drive test in the city of Berlin. Our drive test route contains
different geographical areas, including a park, a highway, and a tunnel in order to get a very diverse data set that is representative of real-world use. By using our own full-stack sidelink implementation \cite{lindstedt2020open}, we could gather data with very high time resolution. Our data set has been made publicly available\footnote{The data can be obtained at \url{https://github.com/fraunhoferhhi/sidelink-mcs-measurements}.} in order to facilitate further research on \gls{ml} methods for sidelink communication.

One important aspect of the Mode 4 specifications is that there is no feedback channel that can adapt transmission parameters based on the channel link quality \cite{molina2018configuration}. In many research studies, the authors have tried to add adaptation capability to C-V2X without changing the protocol. For example, vehicle location and speed were used as side-channel information for optimizing the communications \cite{ku2022adaptive}. 
In another study, the authors propose to adapt transmission power and control message intervals in highly dense environments with severe collisions and interference, to improve the \gls{prr} \cite{kang2021atomic}. In a subsequent study, the authors  demonstrate a multi-agent deep reinforcement learning-based algorithm in sensing-based \gls{sps} to select resources more intelligently and to reduce packet collisions over unmodified  \gls{v2x} Mode 4 \cite{gu2022multiagent}.

The main contributions of our paper are mainly in two areas. Firstly, we have generated a real-world data set by extensive drive tests  through a set of different geographical locations and also made this data set public for further research. Secondly, we have ventured into the issue of fixed transmission parameters in 3GPP \gls{v2x} ( Mode 4 ) Release 14 by employing a machine learning based \gls{mcs} adaption, which permits more flexible data transmission in varying channel conditions. Eventually, in the case of LTE sidelink, we achieve significantly higher goodputs than the best possible non-adaptive method.  

We organized the rest of this paper as follows. First, \cref{drive_test_data_set}
describes the measurement setup and the data set we acquired for our evaluation. Next, \cref{evaluation_methodology} presents the algorithms we used for maximizing goodput. Then, in \cref{prediction_results_section}, we compare the prediction performance of different algorithms. Finally, in \cref{conclusions}, we highlight the main findings.

\section{Drive test data set} \label{drive_test_data_set}

In the following, we describe the hardware setup which we used to perform the measurements, followed by a presentation of the data processing steps and some statistical properties of the data set.

\begin{figure}
\centering
\includegraphics[width=1.0\linewidth]{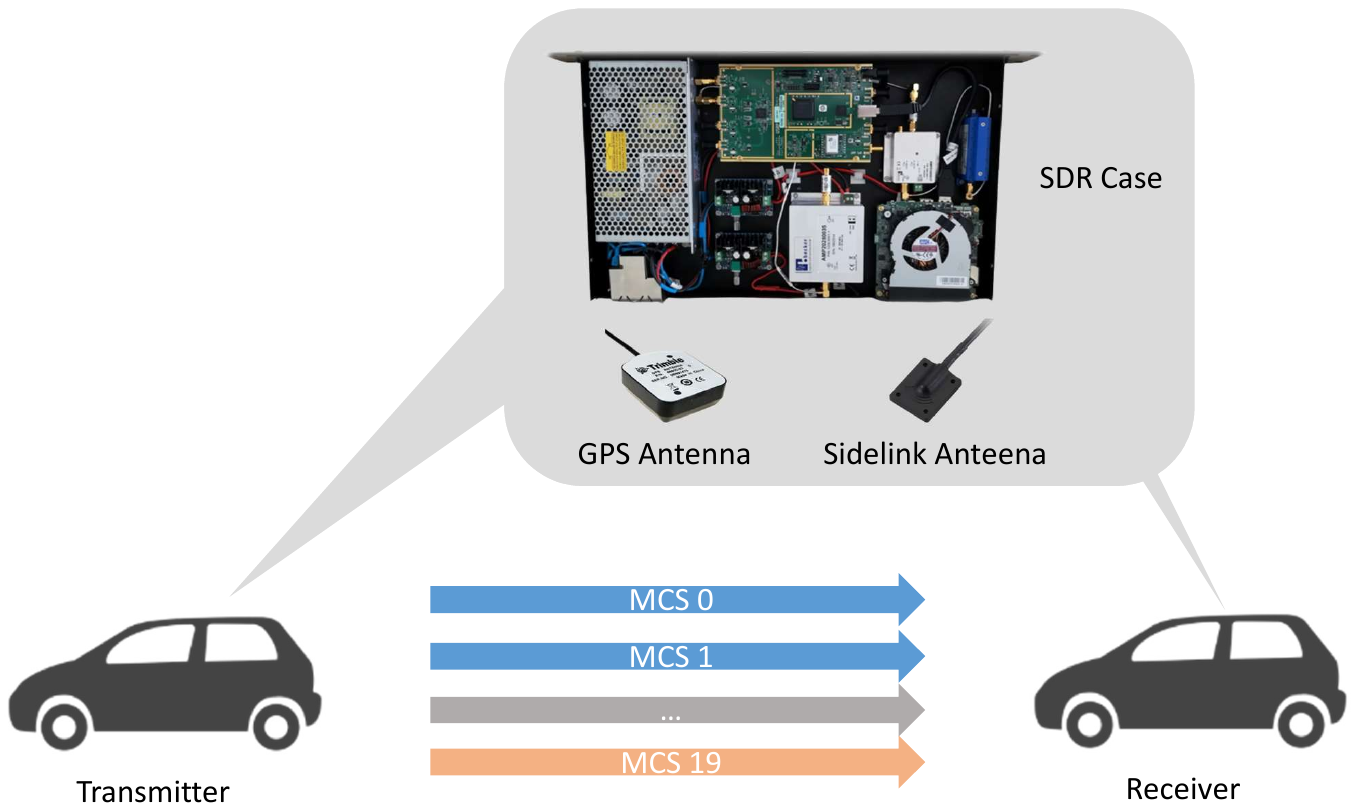}
\caption{Schematic description of measurement setup \cite{lindstedt2020open}.} 
\label{fig:sdr_case}
\end{figure}

For sidelink communication, we used a \gls{sdr} sidelink unit that was previously developed in \cite{lindstedt2020open}. The general setup is shown schematically in \cref{fig:sdr_case}. The sidelink case is equipped with a sidelink antenna and a \gls{gps} antenna, both of which we attached to the roof of the car. Two cars were equipped with one sidelink unit each. Transmission was done with a center frequency of \SI{5.9}{GHz} and a transmit power of \SI{17.3}{dBm}. Both sidelink units were synchronized using \gls{gps} timestamps.

Usually, the sidelink standard prevents sending data every millisecond but requires larger intervals to also give other users transmission opportunities. However, in order to collect enough data, we modified the sidelink Mode 4 software from \cite{lindstedt2020open} in order to quickly sweep over all available \gls{mcs} levels (0 to 19). Moreover, we use one sidelink unit as a dedicated transmitter and the second one as dedicated receiver. We disabled any sensing-based scheduling algorithms in order to use all the 48 \gls{rb} for transmission while sweeping over \gls{mcs} levels. More specifically, a single large packet whose size is chosen to fully utilize the available throughput is transmitted for \SI{1}{ms} with a specific \gls{mcs} level,  after which we switch to the next \gls{mcs} level. The second sidelink unit acts as a dedicated receiver and writes every successfully decoded data packet into a \gls{pcap} based trace for offline analysis. Additionally, physical layer measurements like \gls{snr}, \gls{rsrp}, and \gls{rssi} are also saved to the \gls{pcap} file.

For processing, we parse the information about the successfully decoded packets from the \gls{pcap} file. Due to the fact that one packet was sent per millisecond, we can reconstruct which packets could not be successfully decoded. The physical layer measurements for the missing packets are then reconstructed using linear interpolation from the surrounding packets. Finally, we aggregate each \gls{mcs} sweep and save the highest \gls{mcs} level that could be successfully decoded, while taking the mean of the physical layer measurements. This highest usable \gls{mcs} level can then be used as a target for the prediction algorithms. Context information from \gls{gps} is then merged according to the timestamp.

\begin{figure}
\centering
\includegraphics[width=1\linewidth]{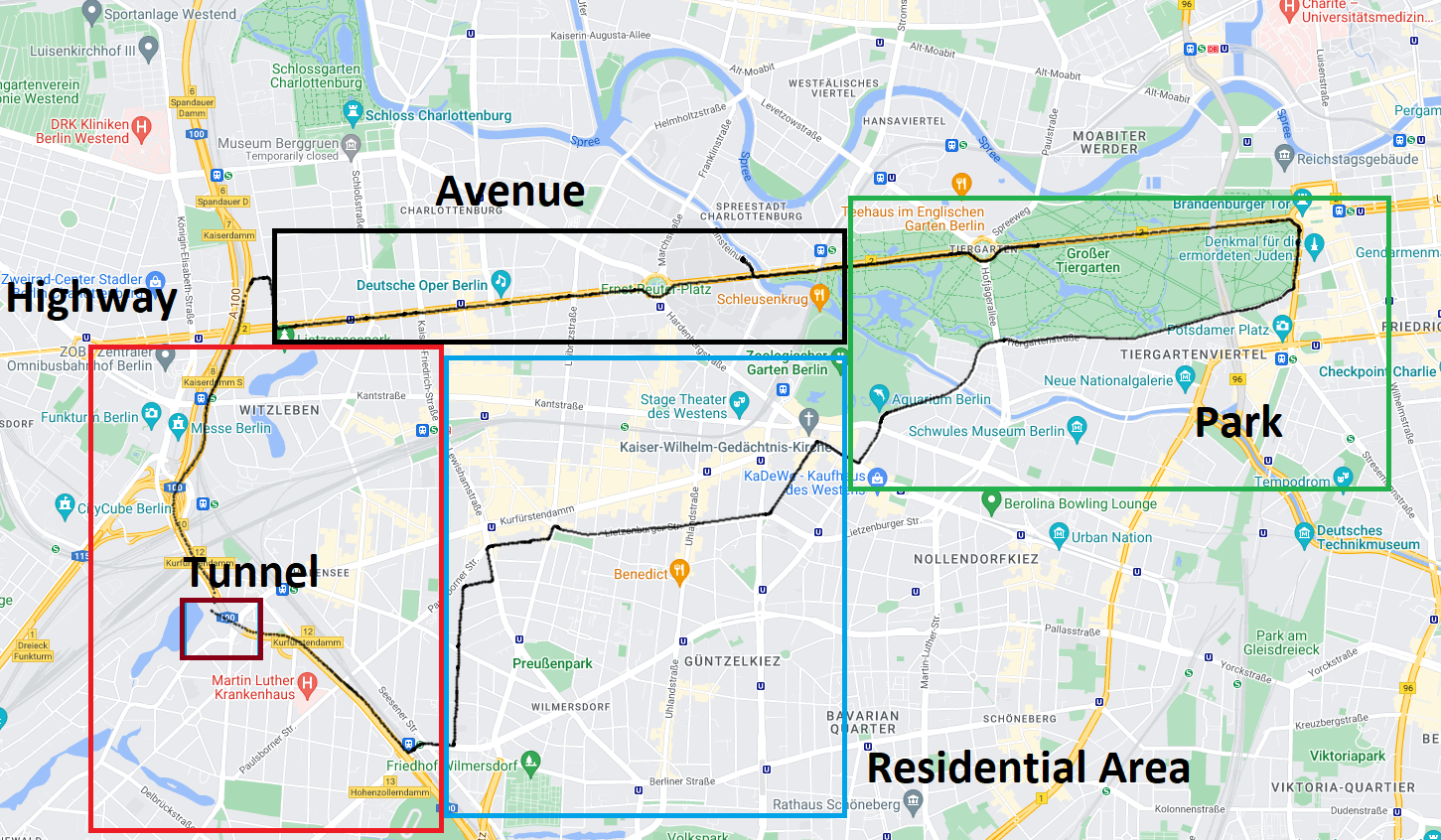}
\caption{Test Drive Route (©2023 GeoBasis-DE/BKG (©2009), Google).}
\label{fig:test drive route}
\end{figure}

In order to gather data in a diverse set of conditions, we devised the route shown in \cref{fig:test drive route}, which contains an avenue, a park, a highway, a residential area, and a tunnel in Berlin. Data was collected from two cars driven in close proximity for a total of six rounds during the course of one day. In total, we have collected 608542 samples. 

Information was collected from a total of two sources, namely physical layer measurements from the \gls{ue} (named \emph{Base}), and positional information from \gls{gps} (named \emph{Positions}). \Cref{tab:features} shows the list of all available features.  

\begin{table}[h]
\caption{Overview of available features.}
\label{tab:features}
\begin{center}
\begin{tabular}{p{10mm}p{10mm}p{50mm}}

\toprule
\textbf{Group} & \textbf{Source} & \textbf{Features}\\

\midrule
Base & UE & \Acrfull{snr}, \acrfull{rsrp}, \acrfull{rssi}, Noise power, Rx power\\
\addlinespace
Positions & GPS & Latitude, Longitude, Velocity\\
\addlinespace
\bottomrule

\end{tabular}
\end{center}
\end{table}

\begin{figure}[h]
    \centering
    \includegraphics[width=1\linewidth]{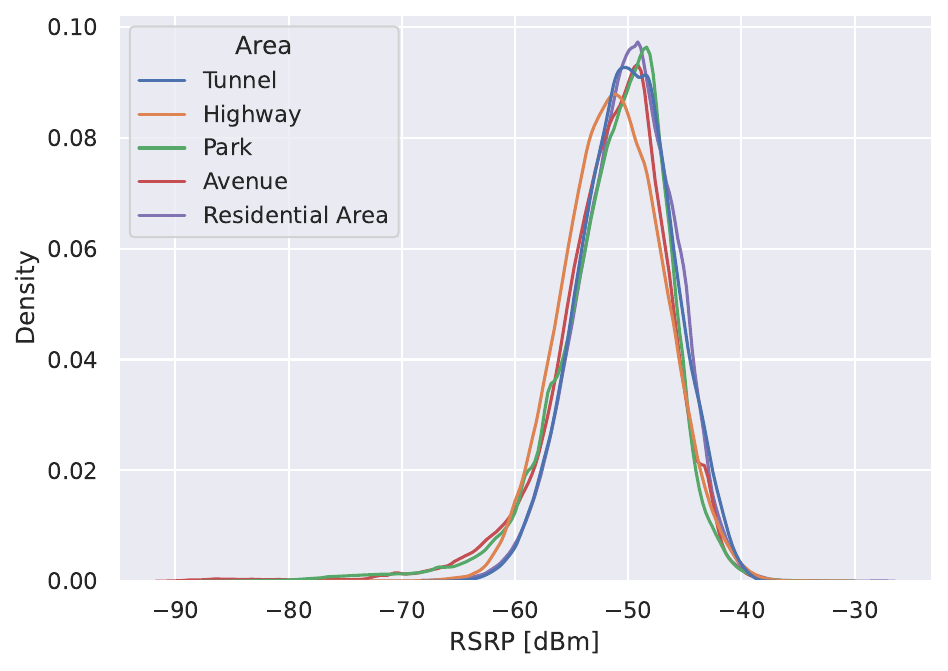}
    \caption{\Gls{kde} of \gls{rsrp} of different geographical areas.}
    \label{fig:kde_of_rsrp}
\end{figure}

\begin{figure}[h]
    \centering
    \includegraphics[width=1\linewidth]{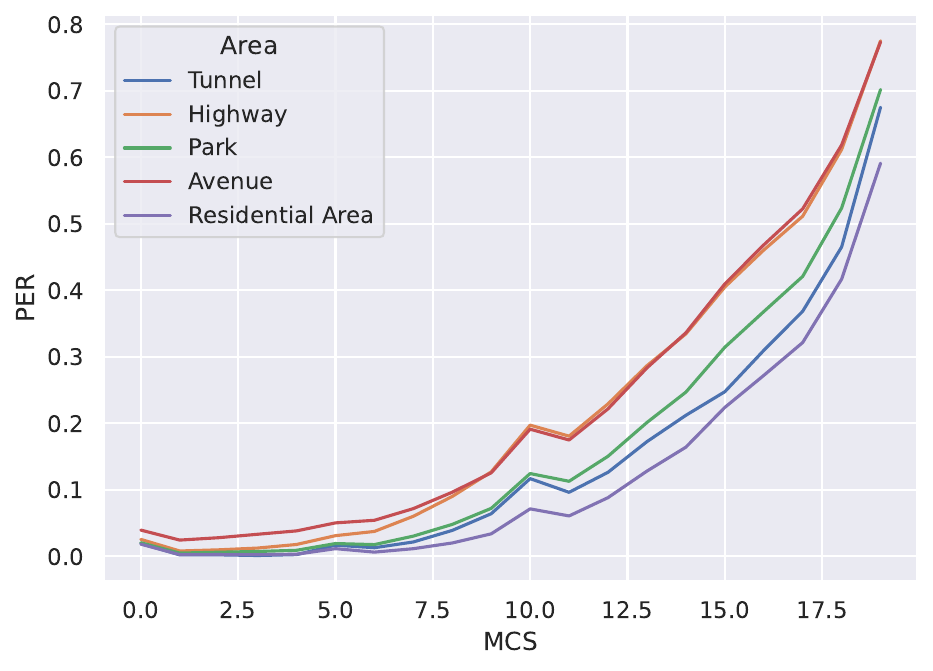}
    \caption{\Gls{per} of different \gls{mcs} levels of different geographical area.}
    \label{fig:per_over_mcs_level}
\end{figure}

\begin{figure}
    \centering
    \includegraphics[width=1\linewidth]{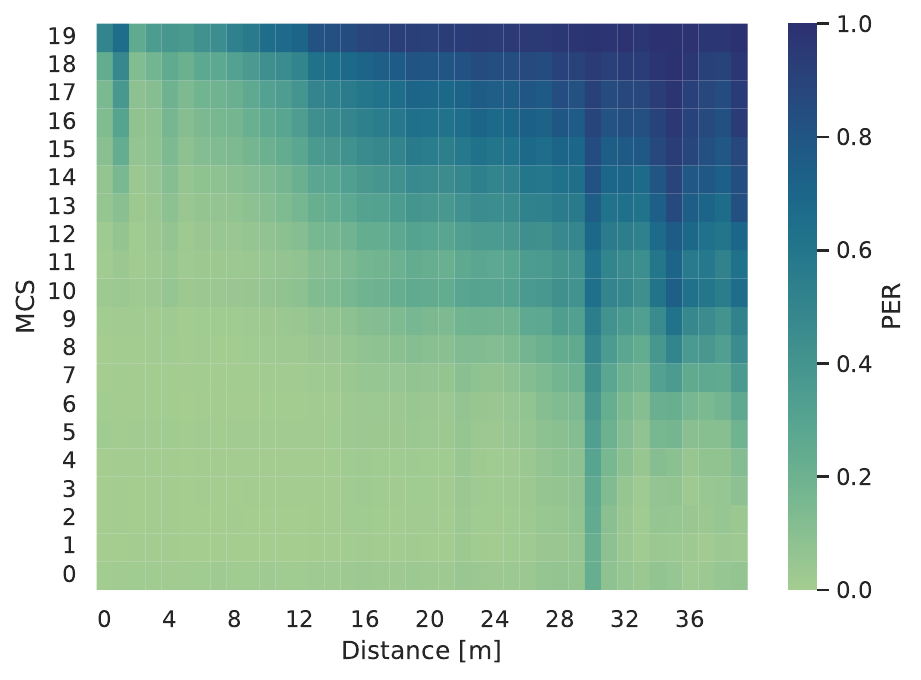}
    \caption{\Gls{per} for different \gls{mcs} levels and distances.}
    \label{fig:distance_vs_mcs}
\end{figure}

Subsequently, we summarize a statistical analysis of the collected data set.
In \cref{fig:kde_of_rsrp}, we show the continuous probability density of the \gls{rsrp}, where the different areas refer to the ones shown in \cref{fig:test drive route}. The \gls{per} for different \gls{mcs} levels split by area is shown in \cref{fig:per_over_mcs_level}\footnote{The reason that \gls{mcs} 11 has lower \gls{per} than \gls{mcs} 10 is that both \gls{mcs} levels support the same data rate, but \gls{mcs} 11 uses 16-QAM as opposed to QPSK, while having a lower code rate, thus leading to lower \gls{per}.}. The reason that the residential area, tunnel, and park have generally lower \glspl{per} than avenue and highway can be attributed to lower distances, which are due to the lower average driving speeds in these geographical areas. In order to further evaluate the influence of distance on \gls{per}, we show the \gls{per} for different distances and \gls{mcs} levels in \cref{fig:distance_vs_mcs}. The (inverse) correlation between distance and \gls{per} can be clearly seen.

\section{Machine Learning Approach}
\label{evaluation_methodology}

In the following, we present the machine learning algorithms that we used to predict the optimal \gls{mcs} level, followed by information on hyperparameter optimization and on the general evaluation methodology.

We evaluated a total of four algorithms, \gls{rf} \cite{breiman2001random, meinshausen2006quantile}, \gls{gb} \cite{chen2016xgboost}, \gls{nn}, and \gls{lr}. The architecture of \gls{nn} contains a feed-forward neural network with the features as input, a number of hidden layers, and the predicted \gls{mcs} level as a single output, using a fully connected topology. The hidden layers employ an activation function, while L1/L2 regularization is used to prevent over-fitting. As an optimizer, we use Adam. The number of hidden layers, number of neurons, and activation functions are treated as hyperparameters that are to be optimized. In addition to these algorithms, as a baseline, we consider the highest goodput that could be achieved by using a fixed \gls{mcs} for all packets. In the context of this study, we consider goodput to be the amount of data that can be successfully decoded at the receiver.

Due to the fact that we try to predict the \gls{mcs} level, our main problem is the inherently asymmetric objective function, i.e., over-predicting the \gls{mcs} level will lead to not decodable packets and thus 0 goodput, while under-predicting the \gls{mcs} level will lead to only a slightly reduced goodput. To deal with this asymmetric nature, we use quantile prediction, which we implemented by using the pinball loss function for \gls{nn} and \gls{gb} and by using quantile regression forests \cite{meinshausen2006quantile} instead of regular \glspl{rf}. The pinball loss function is defined as
\begin{equation}
    \mathcal{L}_\tau(y, \hat{y}) = \left\{ \begin{array}{rcl}
\tau(y - \hat{y}) & \mbox{if} & y \geq \hat{y} \\ 

(1-\tau)(\hat{y} - y) & \mbox{if} & y < \hat{y}, \\
\end{array}\right. \label{eqn:pinball_loss}
\end{equation}
where $\tau$ is the desired quantile and $y$ and $\hat{y}$ are the measured and predicted \gls{mcs} levels, respectively \cite{wang2019probabilistic}. The desired quantile $\tau$ will be treated as a hyper-parameter that is to be optimized. Minimizing pinball loss with \gls{lr} is usually done by means of 
solving a linear programming problem \cite{koenker2005quantile}. However, this is not possible for large data sets due to prohibitive computational complexity. For this reason, the results in this paper for \gls{lr} generally refer to the results derived via the ordinary least squares algorithm, which minimizes the \gls{mse}. When the results explicitly refer to quantile regression, we derived these results by using \gls{sgd}.

In pursuit of hyperparameter optimization, we apply a randomized search over a suitably defined probability distribution for each parameter. We chose a randomized search where we employed the leave one group out method as our cross-validation technique. Leaving one group out uses one round as test data, and the remaining ones as training data, until each unique round has served as a test set once.
This comprehensive hyperparameter optimization method bolsters the robustness and generalization of the model to perform reliably across different real-world environments. By employing this approach, we get a set of results and their corresponding hyperparameters. From them, we considered the hyperparameters that gave the best result for \gls{rf}, \gls{gb}, and \gls{nn} with 100 iterations. Due to the cost of computation, we ran hyperparameter optimization once per algorithm.\footnote{The optimized values for the hyperparameters are documented in the GitHub repository.}

\begin{figure}[h]
    \centering
    \includegraphics[scale=0.55]{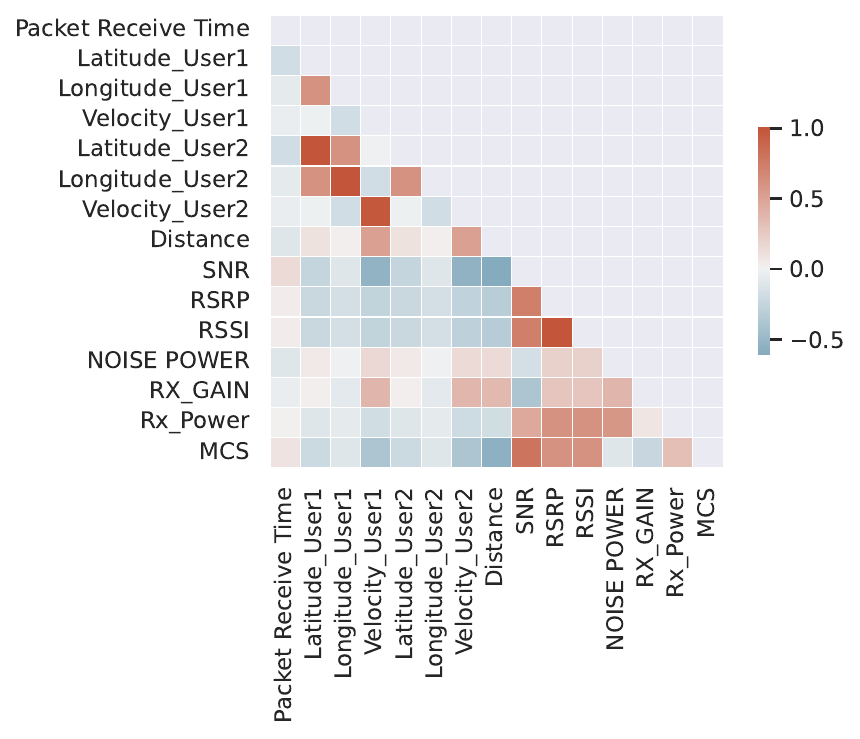}
    \caption{Pearson correlation coefficients between different features and \gls{mcs}. }
    \label{fig:correlation}
\end{figure}

What remains is to determine the optimal set of features. In order to do so, we attempt to order the features by importance. A simple approach to do so is to calculate the correlation between the features and the target. The result is shown in \cref{fig:correlation}, where it can be seen that \gls{mcs} level has the strongest correlation to the features that are related to signal strength (i.e., \gls{snr}, \gls{rsrp}, \gls{rssi} and receive power). However, we note that this approach has serious limitations as it only captures linear dependencies not casual dependencies \cite{correlation}.

In order to better capture the impact that the individual features have on the quality of prediction, we additionally employ the more powerful permutation feature importance method \cite{breiman2001random}. Permutation feature importance measures the decrease in model performance when the values for a specific feature are randomly shuffled. Using this method, we determined that the features sorted by importance are SNR, Rx power, RSSI, distance, RSRP, noise power, speed (of user 2 and 1), latitude (of user 2 and 1), longitude (of user 2 and 1), and Rx gain.

\section{Prediction results}
\label{prediction_results_section}

In order to evaluate the performance of the algorithms defined above, we used one round of the drive data as a test set and the remaining rounds as a training set. We repeated this process for each round and took the average of all rounds. As a performance metric, we calculated the throughput by multiplying the transport block size for the (rounded) predicted \gls{mcs} level and 48 \glspl{rb} according to the 3GPP standard \cite[Table 8.6.1-1, 7.1.7.2.1-1]{etsi2016136} with 1000 transport blocks per second. We then considered this throughput to be goodput, unless the predicted \gls{mcs} level was larger than the highest usable \gls{mcs} level, in which case we set the achievable goodput to 0 for this sample.

\begin{figure}[h]
    \centering
    \includegraphics[scale=.5]{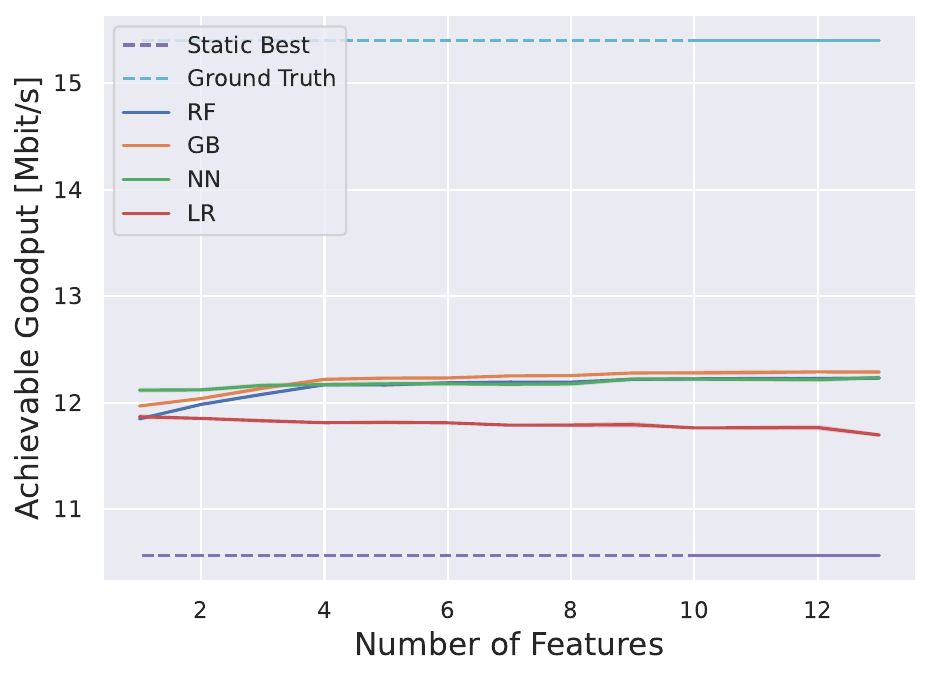}
    \caption{Achievable goodput for different number of features.}
    \label{fig:Feature Combination}
\end{figure}

First, we looked at the influence of the number of features that are used for prediction on the achievable goodput. In order to do so, we sorted the features based on importance according to the method introduced above, and trained the models using only the $N$ most important features. As shown in \cref{fig:Feature Combination}, it can be clearly seen that the performance increases with an increasing number of features for all algorithms except linear regression. However, the performance increase from using more than four features is very small, so in a practical implementation, where computational complexity is a restricting factor, only the four most important features should be used.

\begin{figure}[h]
\centering
\includegraphics[scale=0.5]{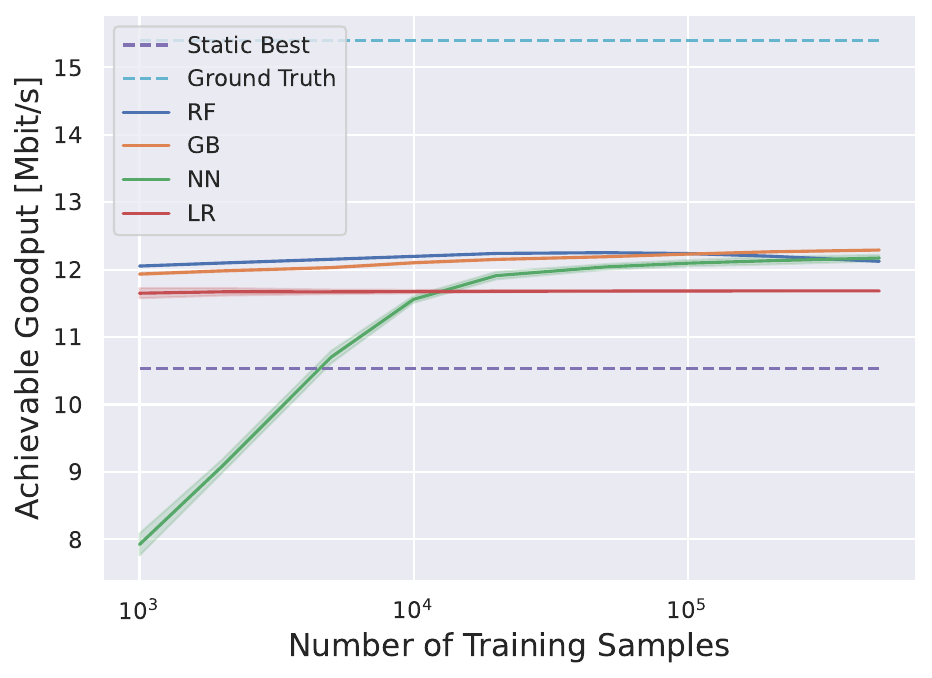}
\caption{Predicted goodput for a varying number of training samples. Shaded regions show the standard deviation of the mean.}
\label{fig:num_training_samples}
\end{figure}

Next, we look at the performance of the different algorithms when only a reduced number of training samples is available. As shown in \cref{fig:num_training_samples}, neural networks need a much larger number of training samples to achieve satisfactory performance, while for the other algorithms increasing the number of training samples only gives a small improvement in performance. This behavior is consistent with the literature \cite{xu2021deep}.

Since using all available training samples and features provides the best results, we use this configuration in a final performance assessment. As shown in \cref{tab:prediction_results_for_diff_loss_function}, the algorithm with the best performance is \gls{gb}, while \gls{nn} and \gls{rf} show similar performance. For comparison, the ground truth gives a goodput of \SI{15.397}{Mbit/s}, while using the best static \gls{mcs} level gives a goodput of \SI{10.541}{Mbit/s}. In order to assess the effectiveness of our quantile regression approach, we compare our results to conventional regression algorithms. This comparison shows that by using quantile regression the goodput is increased by about \SI{0.5}{Mbit/s} over using \gls{mse} loss as loss function and by about \SI{0.8}{Mbit/s} over using \gls{mae} loss, regardless of the algorithm, thus demonstrating the effectiveness of using quantile regression.

\begin{table}[h]
\caption{Prediction results based on different loss functions.}
\label{tab:prediction_results_for_diff_loss_function}
\begin{center}
\begin{tabular}{llll}

\toprule
& \multicolumn{3}{c}{Achievable Goodput [\si{Mbit/s}]}\\
\cmidrule{2-4}

Algorithm & Quant. Reg. & \gls{mse} & \gls{mae}\\

\midrule
Gradient Boosting & \textbf{\num{12.295}} & \num{11.880} & \num{11.520}\\
Neural Network & \num{12.263} & \num{11.818} & \num{11.490}\\
Random Forest & \num{12.232} & \num{11.794} & \num{11.456}\\
Linear Regression & \num{12.033} & \num{11.736} & \num{11.549} \\

\bottomrule

\end{tabular}
\end{center}
\end{table}

\section{Conclusions}
\label{conclusions}
    
In this paper, we describe the measurement setup and the resulting data set of an extensive sidelink measurement campaign in a realistic urban setting. Based on the collected data, we train machine learning algorithms to optimize the \gls{mcs} to maximize the goodput. We demonstrated that significantly higher goodput can be achieved by the machine-learning aided dynamic selection than those achieved by fixed configuration policies, and we additionally show the benefits including vehicle location, velocity, and distance to other vehicles as features to predict \gls{mcs}. Furthermore, our approach does not require any feedback information or any modifications to the current protocol.

The proposed approach dynamically changes the \gls{mcs} and our results show more than \SI{16}{\%} gain in goodput compared to the theoretically best possible result achievable using fixed transmission configuration policies in our real-world data set. In practice, we expect the achieved gains to be even higher, since it is highly difficult to select in advance the \gls{mcs} that will provide the best results. In real-world scenarios, the environment changes rapidly, due to changing distances, the presence of physical obstruction, and interference, and these factors contribute to frequent channel condition fluctuation.

The collected data set has been made fully available in order to be used not only in \gls{mcs} prediction (for example, by evaluating the performance of a reinforcement learning agent for the prediction task), but also for other \gls{ml}-aided optimization research.

\addtolength{\textheight}{-12cm}   


\bibliographystyle{IEEEtran}  
\bibliography{aipsamp.bib}
\vspace{12pt}

\end{document}